\pdfoutput=1

\documentclass[11pt]{article}

\usepackage[final]{acl}

\usepackage{times}
\usepackage{latexsym}
\usepackage{comment}
\usepackage{booktabs}
\usepackage{hyperref}
\usepackage[T1]{fontenc}

\usepackage[utf8]{inputenc}

\usepackage{microtype}

\usepackage{inconsolata}
\usepackage{enumitem}

\title{Instructions for *ACL Proceedings}

\usepackage{inconsolata}
\usepackage{comment}
\usepackage{graphicx}       
\usepackage{multirow}
\usepackage[draft,textsize=footnotesize,textwidth=15mm]{todonotes}
\usepackage{color}
\usepackage{soul}
\usepackage{amsmath}
\usepackage{algorithm} 
\usepackage{algpseudocode}
\usepackage{tcolorbox}
\usepackage{enumitem}
\usepackage{pifont}
\usepackage{setspace}

\title{DCS: Dynamic Corrective Self-Distillation for Better Fine-Tuning of Pretrained Models}

 \author{
 \textbf{Ibtihel Amara}$^{1}$,\quad \textbf{Vinija Jain}\textsuperscript{2,3\dag}\quad\textbf{Aman Chadha}\textsuperscript{2,3\dag} \quad\\
 $^{1}$McGill University\quad $^{2}$Stanford University \quad  
$^{3}$Amazon AI  \\}

\begin{document}
\maketitle
\renewcommand{\thefootnote}{\fnsymbol{footnote}}
\footnotetext[2]{Work does not relate to position at Amazon.}
\renewcommand*{\thefootnote}{\arabic{footnote}}
\setcounter{footnote}{0}
{\makeatletter\acl@finalcopytrue
  \maketitle
}
\begin{abstract}
 We tackle the challenging issue of aggressive fine-tuning encountered during the process of transfer learning of pre-trained language models (PLMs) with limited labeled downstream data. This problem primarily results in a decline in performance on the subsequent task. Inspired by the adaptive boosting method in traditional machine learning, we present an effective dynamic corrective self-distillation (DCS) approach to improve the fine-tuning of the PLMs. Our technique involves performing a self-distillation mechanism where, at each iteration, the student model actively adapts and corrects itself by dynamically adjusting the weights assigned to individual data points. This iterative self-correcting process significantly enhances the overall fine-tuning capability of PLMs, leading to improved performance and robustness. We conducted comprehensive evaluations using the GLUE benchmark demonstrating the efficacy of our method in enhancing the fine-tuning process for various PLMs across diverse downstream tasks. 
\end{abstract}

\section{Introduction}
\vspace{-0.5mm}
There has been a remarkable advancement in the field of Natural Language Processing (NLP) in the past few years, thanks to the introduction of pre-trained language models (PLMs). Recent PLMs like BERT \cite{devlin-etal-2019-bert}, RoBERTa \cite{zhuang-etal-2021-robustly}, XLNet \cite{yang2019xlnet12} have revolutionized and shaped the landscape of the field of NLP by demonstrating significant progress across different downstream applications such as machine translation, reading comprehension, and question and answering.
Standard practice involves fine-tuning PLMs directly on labeled data from these tasks. Yet, when faced with limited downstream data, known as aggressive fine-tuning \cite{jiang-etal-2020-smart}, the risk of model overfitting and reduced generalization capacity emerges. Addressing this challenge has spurred various approaches, encompassing hyper-parameter tuning heuristics \cite{howard-ruder-2018-universal, peters-etal-2019-tune}, additional layer integration \cite{houlsby2019parameter, stickland2019bert}, improved training strategies \cite{chen-etal-2020-recall}, and noise-induced fine-tuning methods \cite{ wu-etal-2022-noisytune}. Notably, techniques such as adapters, hypernetworks, LoRA, QLoRA, and GLoRA have gained traction for efficient parameter updates when data is scarce \cite{houlsby2019parameter, hu2021lora,dettmers2023qlora, chavan2023one}.
\begin{figure*}
  \centering
  \vspace{-1mm}  
  \includegraphics[width=0.95\linewidth]{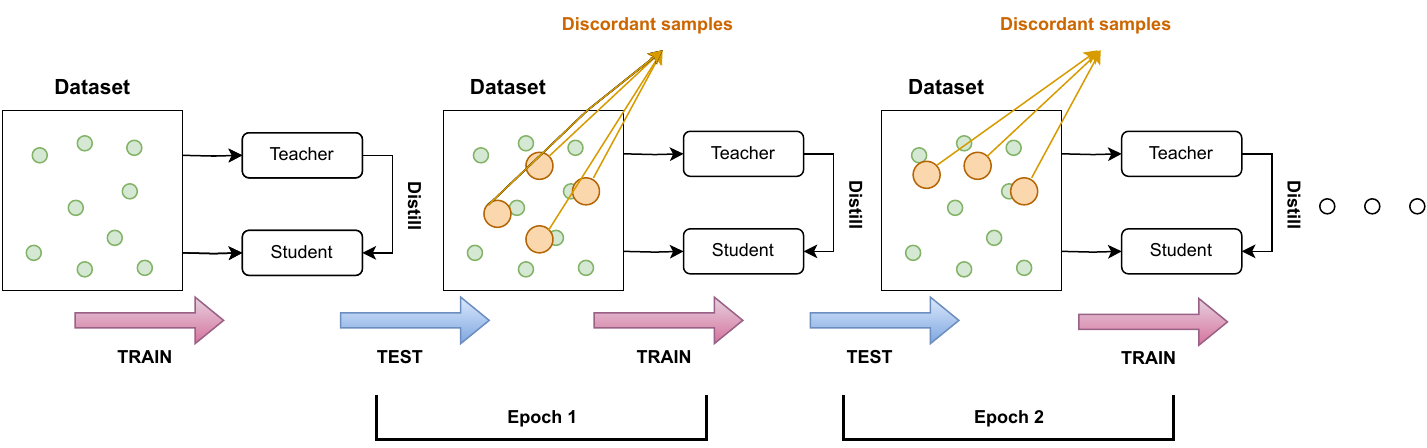}
  \caption{\textbf{Dynamic Corrective Self-Distillation Framework.} DCS iteratively adjusts the weights on the data samples at every epoch. It puts more emphasis on teacher-student discordant (i.e., non-agreeing) samples. We refer to agreement between teacher and student as the level of consensus in their predictions. The distillation component ensures the proper guidance of the student network throughout its optimization process.} 
  \label{fig:fig1}
  \vspace{-1mm}
\end{figure*}
Surprisingly, the potential of distillation as a fine-tuning tool has been somewhat overlooked. Despite its simplicity, distillation offers substantial utility in guiding and optimizing fine-tuning processes. In this work, we introduce Dynamic Corrective Self-distillation (DCS), a straightforward approach inspired by adaptive boosting in machine learning. DCS iteratively adjusts sample weights, prioritizing instances where the teacher and student model disagrees. This emphasis on challenging examples helps improve individual model performance and accuracy in subsequent iterations.
\begin{itemize}
    \vspace{-2mm}
    \item  We propose a flexible and adaptable fine-tuning framework called DCS based on distillation-boosting fusion, which helps large PLMs in avoiding overfitting problems. 
    \vspace{-3mm}
    \item We introduce a self-corrective training framework where the student model is guided by the knowledge of a teacher network.
    \vspace{-3mm}
    \item  Our proposed technique yields a gain in performance of more than 2\% on the GLUE benchmark when compared to vanilla fine-tuning.
    \vspace{-3mm}
\end{itemize}

%
\begin{table*}[t!]
    \centering
    \resizebox{\linewidth}{!}{
    \begin{tabular}{l|cccccccc|c}
    \toprule
    \textbf{Model}                      & \textbf{MNLI}      & \textbf{QQP}   & \textbf{RTE}    & \textbf{QNLI}   & \textbf{MRPC} & \textbf{CoLA} & \textbf{SST}        & \textbf{STS} & \textbf{Avg.} \\ \toprule
    BERT (base) \cite{devlin-etal-2019-bert}             &     84.4        &   90.9    &    67.7       &      91.5  &     87.1      &       58.1    &    93.0       &   89.4  & 82.76                     \\ 
    BERT (base) + DCS  &  \textbf{84.8 }       &   \textbf{91.2}     &    \textbf{71.8}      &   \textbf{ 91.8 }  &      \textbf{87.7}  &       \textbf{59.7}  &      \textbf{93.2}    &      \textbf{89.5} &  \textbf{83.71   }      \\ \midrule
    RoBERTa (base) \cite{wu-etal-2022-noisytune} & 87.5 & 91.7 & 77.1 & 92.7 & 90.1 & 62.9 & \textbf{94.5 }& \textbf{90.8} &  85.91 \\
    RoBERTa (base) + DCS & \textbf{87.6} & \textbf{91.8} & \textbf{81.2} & \textbf{93.0} & \textbf{90.7} & \textbf{63.4 }& 94.2 & \textbf{90.9} & \textbf{86.60 }\\ 
    \midrule
    XLNET \cite{yang2019xlnet12}& 86.6 & 91.2 & 72.9 & 91.6 & 88.1 & 59.6 &  94.4 & 89.6 & 84.25 \\ 
    XLNET + DCS & 86.7&  91.5 & \textbf{74.4} & 91.7  & \textbf{89.2} & 60.5 & \textbf{95.1}  & 88.5 & \textbf{84.70} \\ \midrule
    ELECTRA \cite{clark2020electra} & 88.4 & 91.7 & 75.2& 92.9 & 88.2 & 64.2 & 94.9 & 90.1 & 85.70 \\  
    ELECTRA + DCS &\textbf{ 88.7} &\textbf{ 92.0} & \textbf{84.5} & \textbf{93.5} & \textbf{90.4 }& \textbf{70.4} & \textbf{95.2} &\textbf{91.1} & \textbf{88.22} \\ 
    \bottomrule
    \end{tabular}}
    \caption{Comparison between DCS and Vanilla fine-tuning applied to widely used Pretrained Language Models. The best results are in bold. The results show that DCS leads to substantial improvements across all tasks and among the various PLMs. All DCS values represent the mean values over 3 random seeds.}
    \vspace{-1.5mm}
    \label{tab:my_dev_glue}
\end{table*}

\begin{table}[b!]
    \centering
    \vspace{-3mm}
    \resizebox{\linewidth}{!}{
    \begin{tabular}{l|cccc|c|c}
    \toprule
    \textbf{Methods} & \textbf{CoLA} & \textbf{RTE} & \textbf{MRPC} & \textbf{STS-B} & \textbf{Avg} & $\Delta$\\
    \midrule
    Vanilla FT  & 63.13 & 70.18 & 90.77 & 89.61 &78.42 & 0.00 \\
    \midrule
    Weight Decay & 63.63 &71.99 &90.93 & 89.82 & 79.09 & +0.67 \\
    Top-K FT  & 62.63 &70.90 &91.09 & 89.97 & 78.65 & +0.23 \\
    Mixout & 63.60 &72.12 &91.29 & 89.99 & 79.26 & +0.84 \\
    RecAdam  & 64.33 &71.63 &90.85 & 89.86 & 79.17 & +0.75 \\
    R3F & 64.13 &72.28 &91.18 & 89.61 & 79.30 & +0.88 \\
    CHILD FT(F)  & 63.71 &72.02 &91.22 & 90.18 & 79.29 & +0.87 \\
    CHILD FT(D)  & 64.92 & 73.14 &91.42 & 90.18 & 79.92 & +1.50 \\
    \midrule
    \textbf{DCS (Ours)}  & 64.38 &74.36 & 91.58 & 89.95 & 80.06 & \textbf{+1.65} \\
    \bottomrule
    \end{tabular}}
    \vspace{-0.75mm}
    \caption{Comparison between DCS with other existing fine-tuning methods \cite{daume-iii-2007-frustratingly,houlsby2019parameter, MixoutLee, chen-etal-2020-recall, aghajanyan2021better, xu-etal-2021-raise}. Most values are taken from \cite{xu-etal-2021-raise}. $\Delta$ refers to the performance gain w.r.t VFT. Our findings show DCS performs comparably or even better than other fine-tuning methods.}
    \label{tab:table2}
    \vspace{-1mm}
\end{table}
\vspace{-1mm}
\section{Dynamic Corrective Self-Distillation}
\vspace{-2mm}
To address the issue of limited data for finetuning, we draw inspiration from the adaptive boosting technique in machine learning and propose a dynamic process that iteratively adjusts the sample weights at each epoch of the training process. To achieve accurate weight adjustments, we rely on the knowledge distillation technique \cite{hinton2015distilling}. This technique involves employing a teacher network to guide a student model with the same architecture and capacity. In Figure \ref{fig:fig1}, we provide a visual representation of the DCS framework.
Initially, DCS assigns equal weights to each data point within the downstream dataset. However, as the learning process progresses, DCS modifies these weights. Specifically, greater emphasis is placed on samples where the student and teacher networks diverge in their predictions. In other words, samples that the student network mispredicts compared to the teacher's prediction are assigned higher weights. By incorporating this mechanism, DCS effectively leverages the discrepancy between the student and teacher networks to enhance its learning capability. KD training is performed according to the following objective function.
\vspace{-2mm}
\begin{equation}
\label{eq:total_loss}
    L_{Total} = \alpha L_{CE} + (1- \alpha) L_{KD} 
\end{equation}
\vspace{-1.5mm}
\begin{equation}
\vspace{-1.5mm}
\label{eq:kl_div}
    L_{KD} = -\sum_{i=1}^{N}p(x_i)log(q(x_i))
\vspace{1mm}
\end{equation}
where $p_i$ and $q_i$ are the class probabilities generated by the teacher and student 
\vspace{-1mm}
For the DCS technique, each sample weight is adjusted according to the teacher and student's agreement. For this, the $L_{KD}$ is replaced with the following:
\begin{gather}
\label{eq:weighted_dcs}
L_{KD} = -\sum_{i=1}^{N}w_ip(x_i)log(q(x_i))
\end{gather}
\textls[-10]{such that $w_i = \lambda$ if $\hat{y}_i^T \neq \hat{y}_i^S$, otherwise $w_i = 1$. $\lambda$ is a hyper-parameter higher than 1 $(\lambda >> 1)$. $\hat{y}_i^T$ and $\hat{y}_i^S$ are the predictions of the teacher and the student, respectively, for a sample point $x_i$. 
We provide the steps for performing DCS in Algorithm \ref{app:algo}.}\\
\noindent\underline{\textbf{How can self-KD help finetune the student?}} Recent studies by Furlanello et al. \cite{furlanello2018born} and Stanton et al. \cite{stanton2021does} have highlighted the potential of self-distillation, where the students created through this process can actually surpass the performance of their teachers. This intriguing outcome stems from the fact that distillation inherently involves simplifying the student's model based on the teacher's knowledge. The crucial point here is that it doesn't need an exact match between the student and the teacher. In fact, striving for a perfect match would hinder the student's ability to learn and outperform the teacher. These research works \cite{stanton2021does,furlanello2018born} study the significance of not merely mimicking the teacher but rather distilling and imparting vital knowledge in a manner that amplifies the student's performance.\\
\noindent\textbf{\underline{Weighing discordant samples.}} Introducing sample weighting has been demonstrated to enhance the KD process \cite{lu2021rw}. More specifically, DCS assigns increased weights to the discordant samples, amplifying their impact in the subsequent stages of the student model training. This technique steers the student model toward challenging samples by leveraging the teacher's soft labels while concurrently adhering to the guidance provided by the hard labels (i.e., the ground truth labels).
\vspace{-1mm}
\section{Experiments and Results}
\vspace{-1mm}
\subsection{Experimental Setup}
 \vspace{-0.5mm}
We conduct our experiments on various datasets from the GLUE benchmark \cite{wang2018glue}, a set of datasets containing different tasks such as natural language inference, sentiment analysis, and sentence similarity. Following previous research works, we fine-tune our pretrained models on the training set and directly report results on the dev set. We use the pretrained models and codes provided by HuggingFace \cite{wolf-etal-2020-transformers}. Appendix \ref{app:hpt} elaborates on our hyperparameter sweep setup.
\vspace{-1.5mm}
\subsection{Results}
\vspace{-1.0mm}
\underline{\textbf{Comparison to Vanilla Fine-Tuning.}}
We present compelling results of DCS on eight GLUE tasks using four commonly used PLMs. Following the evaluation methodology of related works \cite{xu-etal-2021-raise,wu-etal-2022-noisytune}, we compare the performance of DCS against vanilla fine-tuning. Table \ref{tab:my_dev_glue} clearly demonstrates that DCS substantially outperforms vanilla fine-tuning across all tasks, affirming its superiority and potential for enhancing PLM fine-tuning. On average, DCS achieves an approximate 1\% improvement across all GLUE benchmark tasks using the BERT base model, and a substantial 2.5\% average score increase on ELECTRA.
Notably, DCS shows significant performance gains on smaller downstream tasks as well. The RTE dataset, the smallest in the GLUE benchmark with only 2.5K samples, exhibits a remarkable boost with DCS. Across various PLMs, we observe notable increases, such as nearly 8\% with ELECTRA and approximately 4\% with RoBERTa and BERT for RTE.
\\
\underline{\textbf{Comparison to Existing FT Methods.}}
\textls[-6]{In this section, we assess the performance and the effectiveness of our proposed fine-tuning technique DCS. For this, we follow the work in \cite{xu-etal-2021-raise} and proceed to compare prior methods on the BERT-large model and report the mean values across different seeds. Results of this comparison are shown in Table \ref{tab:table2}. We observe that all fine-tuning techniques improve upon vanilla fine-tuning. Our technique, which is considered a simple and straightforward technique compared to the existing methods, is as good as or even substantially outperforms these fine-tuning schemes. Mainly we observe a similar trend as in Table \ref{tab:my_dev_glue}, where DCS depicts a significant increase on RTE (the smallest dataset within the GLUE benchmark).}
\begin{figure}[h]
    \centering
    \vspace{-4mm}
\includegraphics[width=\linewidth]{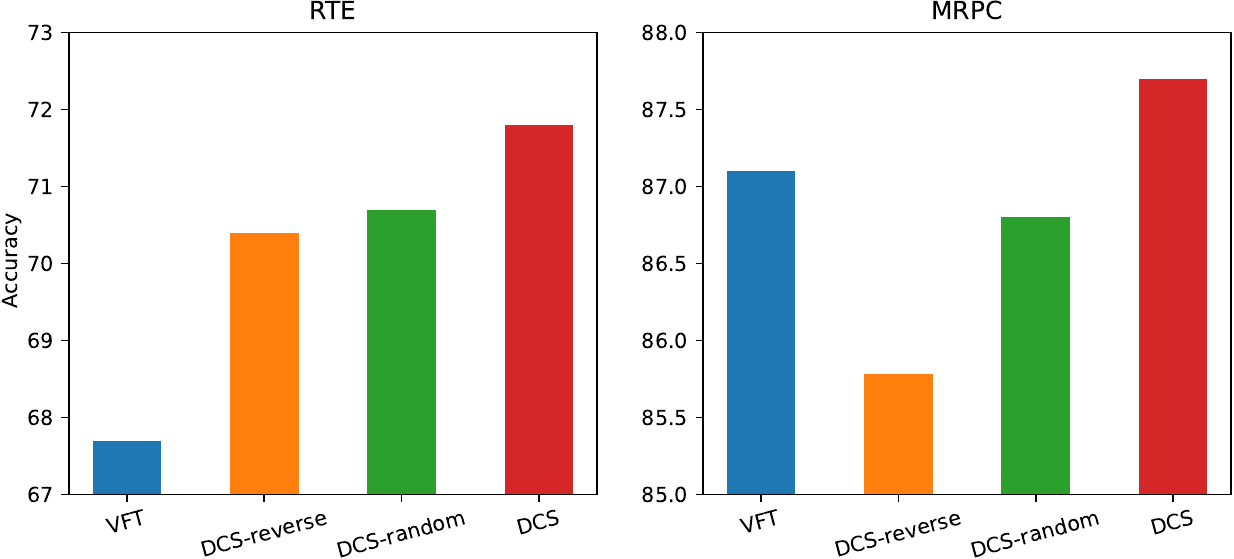}
    \caption{Effect of different weighting strategies of DCS on the performance. Results show that leveraging the discordant teacher-student samples benefits our DCS fine-tuning framework.}
    \label{fig:weight_strategy}
    \vspace{-1mm}
\end{figure}
\vspace{-0.5mm}
\subsection{Ablation and Analyses}
\begin{table}[h]
\vspace{-1.5mm}
    \centering
    \resizebox{\linewidth}{!}{
    \begin{tabular}{l|ccc}
    \toprule
    \textbf{Model and Technique} & \textbf{RTE} & \textbf{CoLA} & \textbf{MRPC} \\ \midrule
    BERT DCS & 71.11 & 59.61 & 87.58  \\
    BERT DCS (w/o weighting) & 70.39 & 58.38 & 87.25 \\
    \midrule
    ELECTRA DCS &  84.47 & 70.57 & 90.49\\
    ELECTRA DCS (w/o weighting) & 83.75 & 70.06 & 90.19\\
    \bottomrule
    \end{tabular}}
    \vspace{-1mm}
    \caption{\textls[-10]{\textbf{Ablation study of DCS.} Without weighting involves distillation without the dynamic self-correction component. Removing both weighting and distillation components leads to vanilla fine-tuning. Results emphasize the importance of combining both components for enhancing the performance of our fine-tuning framework.}}
    \vspace{-6mm}
    \label{tab:ablation}
\end{table}
\underline{\textbf{Components of DCS.}} The DCS technique comprises two key components that synergistically enhance its effectiveness: (1) The sample re-weighting mechanism, serving as a self-corrective element within DCS, and (2) The offline distillation component, which plays a dual role. Firstly, it identifies samples that require increased attention, and secondly, it continues to guide the student network during the learning process. To assess the significance of these components, we conducted an ablation study, the results of which are presented in Table \ref{tab:ablation}. When removing the self-corrective mechanism (referred to as DCS without weighting), we transition to sole distillation. However, this leads to a noticeable performance decrease compared to DCS. On average, we observe a reduction of more than 1\% in RTE performance for both BERT base (from 71.11\% downto 70.39\%) and ELECTRA (from 84.47\% downto 83.75\%) models. Furthermore, eliminating the distillation component returns us to the vanilla fine-tuning technique. As previously observed, DCS consistently outperforms vanilla fine-tuning, highlighting the superiority of our approach. \\
%
%
\noindent
\underline{\textbf{Leveraging Disagreement.}}
In this section, we investigate the benefits of assigning higher weights to discordant student-teacher samples in DCS and how different weighting strategies impact its performance. Figure \ref{fig:weight_strategy} presents a comparison of various weighting strategies: DCS-reverse, which prioritizes concordant teacher-student samples, and DCS-random, which assigns weights randomly. Analyzing the performance of BERT-base on different datasets, we observe that leveraging the discrepancies between teacher and student networks improves the performance and robustness of DCS. Mainly, DCS-reverse performs the least, suggesting that focusing on concordant samples hampers generalization and performance. Conversely, emphasizing non-agreeing samples between the teacher and student enhances the student's understanding and potential for improved generalization, resulting in performance gains.
\vspace{-2.5mm}
\subsection{Hyperparameter Sensitivity}
\begin{figure}[bt]
    \centering
    \includegraphics[width=\linewidth]{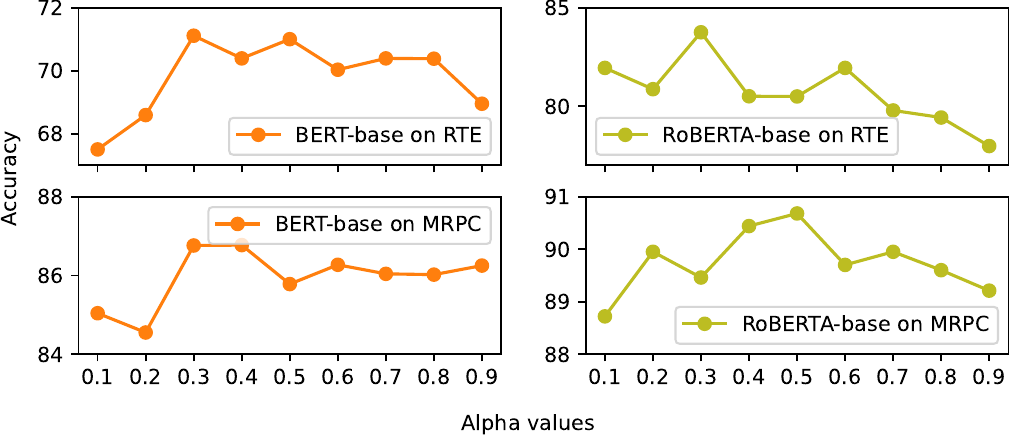}
    \caption{\textls[0]{\textbf{Sensitivity of DCS to $\alpha$.} The results highlight the importance of teacher guidance through distillation. Higher $\alpha$ values indicate a stronger influence on the KD loss, while lower values emphasize the distillation loss. Interestingly, moderate alpha values lead to significant performance improvements, while further increases in alpha show decreasing values.}}
    \label{fig:hp_alpha}
    \vspace{-2.5mm}
\end{figure}

\vspace{-1mm}
\underline{\textbf{Influence of the hyperparameter $\alpha$.}} In KD, $\alpha$ handles the amount of contribution for each of the losses (i.e., cross-entropy and KD). We run different experiments with different values of $\alpha$. As shown in Figure \ref{fig:hp_alpha}, DCS shows higher performance for smaller to intermediate values of $\alpha$, which suggests that the student fine-tuned model benefits from the distillation component of the optimization process. Relevant details can be found in \ref{app:lambda}.
\vspace{-1mm}

\section{Conclusion}
\vspace{-1mm}
In this paper, we introduce DCS, a framework for enhancing the fine-tuning of pre-trained language models (PLMs) on downstream tasks. By incorporating a dynamic self-correction mechanism triggered via self-distillation, DCS achieves remarkable performance gains over vanilla fine-tuning and previous approaches. Through extensive experiments on various tasks from the GLUE benchmark, we demonstrate the superior performance of DCS across four different pre-trained language models. Notably, DCS not only outperforms existing methods but also significantly enhances the generalization ability of the fine-tuned models. One of the main advantages of using our proposed method DCS is flexibility. Unlike other existing methods such as parameter-efficient fine-tuning (PEFT) methods and noise addition, DCS does not limit the flexibility of the model and its ability to adapt to each downstream task.

\section{Limitations}

One main limitation of the DCS method is its computational complexity. DCS involves training a teacher model (of similar size as the student network) using the direct fine-tuning method on the downstream tasks. Then, we use this teacher to better guide a student model using the adaptive corrective approach via knowledge distillation.  However, we argue that teacher models (i.e., vanilla fine-tuned) are abundant, meaning that we can easily find a directly fine-tuned teacher network in public repositories like Hugging face hubs. Therefore, one may not need to consider this as a limitation.

\bibliography{reference}

\newpage
\section*{Frequently Asked Questions (FAQs)}\label{sec:FAQs}

\setlist{leftmargin=1mm}
\begin{itemize}
[leftmargin=4.5mm]
\setlength\itemsep{1.5em}
    \item[\ding{93}] {\fontfamily{lmss} \selectfont \textbf{How do you determine the optimal teacher network to guide the student model effectively during the process of distillation?}}
    \begin{description}
    \item[\ding{224}] 
    The DCS framework does not require a specific type of teacher model or high-performing teacher model. Any pre-fine-tuned teacher model can be used within the DCS framework. In our experiments, we opted for self-distillation, where the teacher and student networks have similar architectures and sizes. The performance of the teacher model does not need to be exceptional or achieve high accuracy. It simply needs to be trained for a few iterations without full convergence. In our experiments, we fine-tuned the teachers for a limited number of epochs (2 epochs). This flexibility in the choice of teacher models allows for broader applicability and ease of implementation within the DCS framework.
    \end{description}

    \item[\ding{93}] {\fontfamily{lmss} \selectfont \textbf{How can we expect the student network to achieve better results on downstream tasks if we choose a teacher network that is not performing well?}}
    \begin{description}
    \item[\ding{224}] 
    Knowledge distillation can still be beneficial for the student network’s optimization process. The main purpose of knowledge distillation is to transfer valuable insights learned by the teacher down to the student \cite{furlanello2018born,stanton2021does}. Although the teacher network did not achieve high performance on its own, it still provides valuable patterns and special relationships in the data. By considering this distilled knowledge, the student has the ability to improve its performance and achieve better results on the downstream task. One main thing to note, is that in our DCS framework, we perform a weighted distillation in which we give higher weights on discordant samples. The latter can be one of the following scenarios: (a) The teacher network correctly classified the sample whereas the student misclassified the sample, or (b) vice versa. In this case, the student network gets to attend to both scenarios and learn from both the good and the bad throughout its learning process, which contributes to its effectiveness and robustness. Furthermore, it is worth mentioning that our student networks receive guidance not solely from the soft labels provided by the teacher but also from the hard labels known as the ground truth labels. This dual guidance alongside the re-weighting strategy serves as an effective regularization strategy.
    \end{description}
    \item[\ding{93}] {\fontfamily{lmss} \selectfont \textbf{What type of knowledge are you using for distilling information from teacher to student? }}
    \begin{description}
    \item[\ding{224}] 
    In DCS, we solely use response-based knowledge. We mainly exploit the logits (i.e., the raw pre-softmax outputs) to align the knowledge between the teacher and the student networks. Our main goal is to propose a straightforward and less complicated framework for fine-tuning. Therefore, we opted for this type of knowledge rather than feature based distillation, which can bring more computational complications to the overall framework. In addition, by aligning the logits, the student model can gain more insights into the finer details of the task guided by the teacher model. 
    \end{description}

    \item[\ding{93}] {\fontfamily{lmss} \selectfont \textbf{DCS is inspired from the adaptive boosting technique in machine learning, which in the end combines multiple weak learners. Do you do something similar?}}
    \begin{description}
    \item[\ding{224}] 
    \textls[0]{For this current version of our work, the same network is being adjusted (i.e., dynamic self-correction) according to the new weighted samples. For inference, we use a single network (i.e., the student network). It is possible to combine multiple versions of our student (i.e. different student model checkpoints) during inference to have a more robust predictions and outcomes.}
    \end{description}

\end{itemize}

\appendix
\renewcommand{\thesubsection}{\Alph{section}.\arabic{subsection}}
\renewcommand{\thesection}{\Alph{section}}
\setcounter{section}{0}
\vspace{-1mm}
\newpage
\section{Appendix}

This section provides supplementary material in the form of additional results, implementation details, etc. to bolster the reader's understanding of the concepts presented in this work.

\subsection{DCS Algorithm}
\label{app:algo}
\begin{algorithm}
\vspace{-1mm}        
	\caption{DCS} 
        \footnotesize{
	\begin{algorithmic}[1]
            \State Load teacher weights $\theta_T$
            \State Initialize student's parameters $\theta_S$
            \For {each epoch} 
            \If {epoch == 0}
            \State \textbf{\textit{Train}} the student according to  Eq.\ref{eq:total_loss}\\ \hspace{\algorithmicindent} \hspace{\algorithmicindent} using the KD loss in Eq. \ref{eq:kl_div}
            \Else
            \State \textbf{\textit{Re-weight}} the samples according to\\
            \hspace{\algorithmicindent} \hspace{\algorithmicindent} the agreement between the teacher\\
            \hspace{\algorithmicindent} \hspace{\algorithmicindent} and the current state of the student
            \State \textbf{\textit{Train}} the student according to  Eq.\ref{eq:total_loss}\\ \hspace{\algorithmicindent} \hspace{\algorithmicindent} using the KD loss in Eq. \ref{eq:weighted_dcs}
            \EndIf
            \EndFor
	\end{algorithmic} 
        }
\vspace{-1mm}        
\end{algorithm}

\textls[0]{The pseudo-code for our DCS framework is outlined in Algorithm 1. A key requirement is to have a pre-trained teacher model with similar architecture and size as the student network. While the algorithm shares similarities with offline knowledge distillation for downstream tasks, the key differentiating factor is the incorporation of the re-weighting mechanism. The latter plays a crucial role in identifying discordant teacher-student samples and assigning them higher weight values, allowing the student network to benefit from the valuable insights provided by these samples. 
It is important to note that, the first epoch of training in DCS is intentionally designed to exclude the re-weighting mechanism. This is based on the assumption that the student network has not yet encountered and learned the complexity and intricacies of the downstream data. By providing this initial epoch as a warm-up phase, the DCS framework enables the student network to gradually adapt to the complexities of the downstream task, leading to improved performance and generalization capabilities.}

\subsection{Influence of the $\lambda$ Hyperparameter}
\label{app:lambda}
\begin{table}[!h]
    \centering
    \resizebox{\linewidth}{!}{
    \begin{tabular}{l|ccccc}
    \toprule
    \textbf{Model $\downarrow$} / \textbf{$\lambda$ $\rightarrow$} &\textbf{1} & \textbf{2} & \textbf{3} & \textbf{4} & \textbf{5} \\ \midrule
    BERT-DCS  & 70.39 & 71.11 & 70.50 & 70.81 &70.63 \\

    \bottomrule
    \end{tabular}}
    \vspace{-1mm}    
    \caption{Influence of the hyper-parameter $\lambda$ on the performance of DCS on the RTE task.}
    \label{tab:lambda}
    \vspace{-2mm}
\end{table}
\vspace{-1mm}
DCS relies on the re-weighting strategy for self-correction. We perform experiments with different lambda values and assess their effect on the performance of DCS. Table \ref{tab:lambda} shows that the optimal value is $\lambda =2 $. Beyond that, we observe a steady outcome on the DCS.

\subsection{Hyperparameter Search Values}
\label{app:hpt}
In this paper, we fine-tune different large pre-trained language models. We performed a simple grid search to find the optimal hyperparameter settings as shown in Table \ref{tab:hptsearch}. 
\vspace{-1mm}
\begin{table}[!h]
    \centering
    \resizebox{1\linewidth}{!}{
    \begin{tabular}{l|l}
    \toprule
    \textbf{Hyperparameters} & \textbf{Search Values} \\ 
    \midrule
    Batch size & \{8, 16\} \\
    Learning rate (lr) & \{1e-5, 2e-5, 1e-6, 2e-6\} \\
    Distillation temperature (T) & 1 \\ 
    $\alpha$ KD & \{0.1, 0.2, 0.3, 0.4, 0.5, 0.6, 0.7, 0.8, 0.9\}  \\
    sample weighting ($\lambda$) & \{2, 3, 4, 5, 6\} \\
    \bottomrule
    \end{tabular}}
    \vspace{-1mm}    
    \caption{\textls[-10]{Hyperparameter search space for all tasks in GLUE.}}
    \label{tab:hptsearch}
    \vspace{-2mm}
\end{table}
\end{document}